\documentclass[10pt,twocolumn,letterpaper]{article}

\usepackage{iccv}              

\usepackage{multirow}
\usepackage{colortbl}
\usepackage{listings}

\definecolor{myred}{RGB}{246,107,77}
\definecolor{mygreen}{RGB}{0,155,85}
\definecolor{myblue}{RGB}{0,130,251}

\definecolor{mygray}{gray}{0.6}
\definecolor{mygray-bg}{gray}{0.9}

\usepackage[scale=0.9]{sourcecodepro}
\lstdefinestyle{mystyle}{
    basicstyle=\ttfamily\small,
}
\definecolor{iccvblue}{rgb}{0.21,0.49,0.74}
\usepackage[pagebackref,breaklinks,colorlinks,allcolors=iccvblue]{hyperref}
\def\paperID{6456} 
\def\confName{ICCV}
\def\confYear{2025}

\newcommand{\datasetname}{3DSRBench}

\title{\datasetname: A Comprehensive 3D Spatial Reasoning Benchmark}

\author{Wufei Ma \quad Haoyu Chen$^\dagger$ \quad Guofeng Zhang \quad Yu-Cheng Chou \\
Jieneng Chen \quad Celso de Melo$^\circ$ \quad Alan Yuille \\
\small Johns Hopkins University \; $^\dagger$Carnegie Mellon University \; $^\circ$DEVCOM Army Research Laboratory
}

\begin{document}
\maketitle
\begin{abstract}

3D spatial reasoning is the ability to analyze and interpret the positions, orientations, and spatial relationships of objects within the 3D space. This allows models to develop a comprehensive understanding of the 3D scene, enabling their applicability to a broader range of areas, such as autonomous navigation, robotics, and AR/VR. While large multi-modal models (LMMs) have achieved remarkable progress in a wide range of image and video understanding tasks, their capabilities to perform 3D spatial reasoning on diverse natural images are less studied. In this work we present the first comprehensive 3D spatial reasoning benchmark, \datasetname, with 2,772 manually annotated visual question-answer pairs across 12 question types. We conduct robust and thorough evaluation of 3D spatial reasoning abilities by balancing data distribution and adopting a novel FlipEval strategy. To further study the robustness of 3D spatial reasoning \wrt camera 3D viewpoints, our \datasetname{} includes two subsets with 3D spatial reasoning questions on paired images with common and uncommon viewpoints. We benchmark a wide range of open-sourced and proprietary LMMs, uncovering their limitations in various aspects of 3D awareness, such as height, orientation, location, and multi-object reasoning, as well as their degraded performance on images from uncommon 6D viewpoints. Our \datasetname{} provide valuable findings and insights about future development of LMMs with strong spatial reasoning abilities. Our project page is available \href{https://3dsrbench.github.io/}{here}.
\end{abstract}    
\section{Introduction} \label{sec:intro}
Recent large multi-modal models (LMMs)~\cite{achiam2023gpt, reid2024gemini, claude} have achieved significant improvements in a wide range of image and video understanding tasks, such as image captioning~\cite{alayrac2022flamingo, li2023blip}, visual question answering~\cite{hudson2019gqa, goyal2017making, liu2023llava, tong2024cambrian}, visual grounding~\cite{zhang2023recognize}, decision making~\cite{lu2024generative, kim2024openvla, brohan2023rt}, and action recognition~\cite{wang2022internvideo, ma2025rethinking}.
Notably, the spatial reasoning ability~\cite{johnson2017clevr, hudson2019gqa, cheng2024spatialrgpt, wang20233d}, \ie, parsing 2D and 3D spatial relationships between objects, serves as a crucial foundation for various high-level reasoning and interaction in downstream tasks. Studying the spatial reasoning ability of current LMMs will help us identify specific types of factual errors, uncover their fundamental limitations, and inform targeted improvements to further advance current LMMs.

Prior datasets~\cite{johnson2017clevr,hudson2019gqa,li2023super,kamath2023s} studying spatial relationships often focused on relationships \wrt the viewer, \eg, object A is to the left of object B from the viewer's perspective. We regard these as 2D spatial relationships as they can be captured merely from 2D bounding boxes of the objects (see \cref{fig:dataset-orientation}). They neglect 3D spatial relationships in the 3D world space or those from an object's perspective. Capturing 3D spatial relationships between objects in the images would help LMMs understand and predict the interactions between objects, and enable a broader range of applications in 3D, \eg, robotics and embodied AI.

\begin{figure*}[t]
  \centering
  \begin{subfigure}{0.515\linewidth}
    \includegraphics[width=\linewidth]{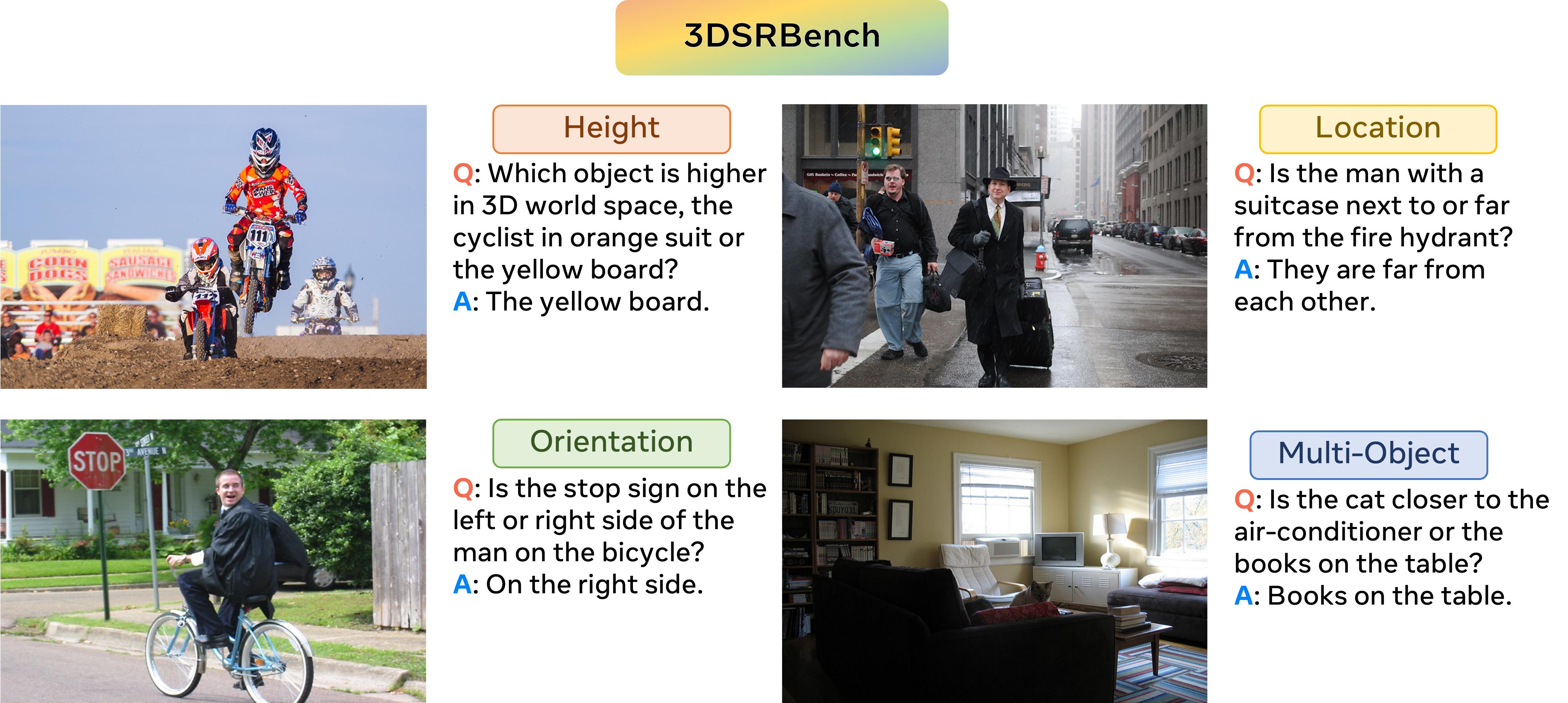}
    \caption{}
    \label{fig:teaser-a}
  \end{subfigure}
  \hfill
  \begin{subfigure}{0.416\linewidth}
    \includegraphics[width=\linewidth]{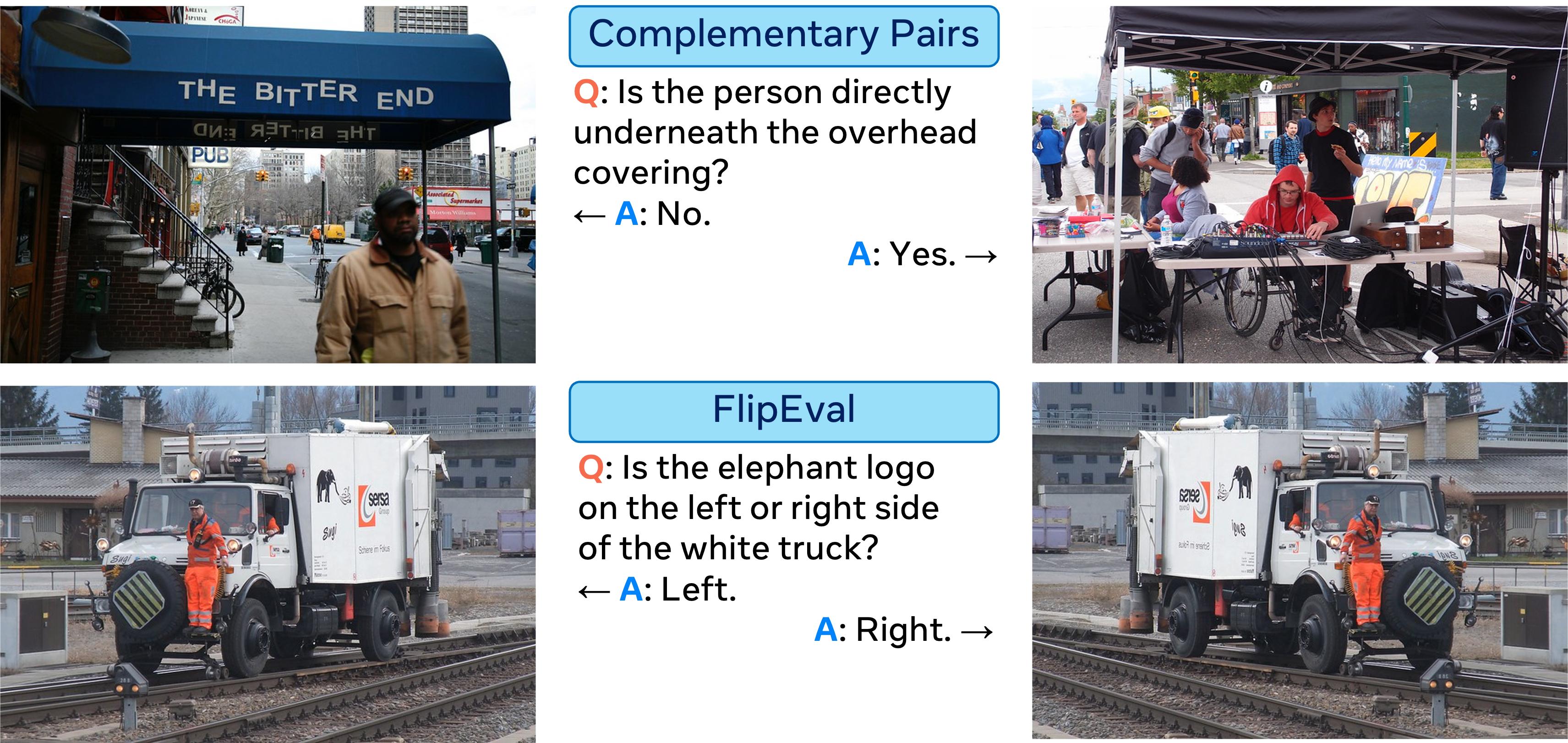}
    \caption{}
    \label{fig:teaser-b}
  \end{subfigure}
  \caption{\textbf{Overview of our \datasetname.} \textbf{(a)} Example questions from the four main types of 3D spatial reasoning questions, \ie, height, location, orientation, and multi-object reasoning. \textbf{(b)} To enable a robust evaluation of the 3D spatial reasoning capabilities, we collect complementary images that lead to opposite answers given the same question and adopt a novel FlipEval strategy to remove left/right biases in 3D with paired VQAs (see \cref{sec:dataset-eval}).}
  \label{fig:teaser}
\end{figure*}

To study how LMMs can capture 3D spatial relationships, previous works often exploited synthetic environments and generated images with 3D ground-truths~\cite{wang20243d,wang2024compositional}.
Visual question-answer pairs were automatically synthesized by applying pre-defined rules to the known 3D scene graphs and object attributes. The synthetic images exhibit a significant domain gap with natural images and lacked the diversity and richness in real-world. More recent works~\cite{cheng2024spatialrgpt} explored real datasets with 3D annotations, \eg, Omni3D~\cite{brazil2023omni3d}. However, images in these datasets are limited to specific domains, such as indoor rooms and self-driving scenes. In general, visual question-answer pairs generated with rule-based methods from 3D annotations (i)~limit the scope of theirs datasets to a small set of rigid object categories, and (ii)~cannot enable a fine-grained and robust evaluation of 3D spatial relationships that can only be achieved with human annotated datasets (see \cref{sec:dataset-design}).

In this work we present the first comprehensive 3D spatial reasoning benchmark, \textit{\datasetname}, that features 2,772 3D spatial reasoning questions from 12 question types on diverse and open-vocabulary entities, including rigid objects, humans, animals, and implicit concepts, such as logo on a car or arrow on a billboard. We manually annotate 2,100 visual question-answer pairs on natural images from the MS-COCO dataset~\cite{lin2014microsoft}, covering 12 subtypes of questions from 4 main categories, \ie, height, location, orientation, and multi-object reasoning. Each category of questions focus on different combinations of 3D properties, such as object 3D location, 3D ground plane, camera extrinsic calibration, and/or object 3D poses. Examples from each question category are presented in Figure~\ref{fig:teaser-a}.


Another challenge of 3D spatial reasoning arises from the 6D viewpoint of the camera, \ie, the 3D location and 3D orientation from which we are viewing the 3D scene. As shown in \cref{fig:hssd}, 3D spatial reasoning questions can be easier for common 6D viewpoints, \eg, ones positioned at the eye level with natural viewing angles, while being more challenging for other uncommon viewpoints. Although uncommon viewpoints are less populated in most image datasets, cameras in embodied AI and robotics are often positioned in these uncommon viewpoints. Hence it is of crucial importance for LMMs to retain good 3D spatial reasoning performance for both common and uncommon viewpoints. To fairly compare the 3D spatial reasoning capabilities of LMMs \wrt different camera viewpoints, we annotate another 672 visual question-answer pairs on multi-view synthetic images rendered from the HSSD dataset~\cite{khanna2024habitat}.

Besides benchmarking a wide variety of open-sourced and proprietary LMMs, our \datasetname{} serves as an important diagnosis benchmark for developing 3D spatially intelligent LMMs. Inspired by previous studies on 3D awareness of visual foundation models~\cite{el2024probing,ma2024imagenet3d}, our \datasetname{} takes one step further and evaluates LMMs on fundamental 3D spatial reasoning questions, which provide valuable insights regarding the 3D awareness of visual encoders~\cite{radford2021learning, chen2024vitamin, he2022masked, oquab2023dinov2, kirillov2023segment} and the 3D reasoning abilities of language models~\cite{chiang2023vicuna, zheng2023judging, touvron2023llama, dubey2024llama}. Such results would shed light on downstream tasks that build on 3D spatial reasoning, such as automatic navigation and robotic manipulation.

To enable a comprehensive and robust evaluation of 3D spatial reasoning abilities, \datasetname{} adopts several key designs: (1) balanced data distributions in multiple aspects, such as balanced answer distribution and complementary images pairs that lead to opposite answers given the same question (see \cref{fig:teaser-b}); (2) avoiding questions with shortcuts or trivial answers; and (3) a novel FlipEval strategy for robust evaluation of 3D spatial reasoning abilities.



Our \datasetname{} significantly advances the evaluation of 3D spatial reasoning abilities and provide valuable findings and insights about the future development of LMMs. We benchmark a wide variety of open-sourced and proprietary LMMs on \datasetname{} and study their 3D spatial reasoning abilities \wrt different types of 3D awareness. We further investigate how various visual encoder designs and scaling of language model sizes can benefit 3D spatial reasoning abilities. Moreover, with the paired images in \datasetname-\lstinline{synthetic}, we analyze the robustness of 3D spatial reasoning abilities \wrt uncommon camera 6D viewpoints. Lastly, by analyzing failure modes of state-of-the-art LMMs, we highlight limitations of current LMMs and discuss possible future improvements. Experimental results on different splits of our \datasetname{} provide valuable findings and insights that will benefit future research on 3D spatially intelligent LMMs.

%

%

\section{Related Works} \label{sec:related}

\paragraph{Spatial reasoning.}
Early works~\cite{johnson2017clevr,hudson2019gqa,li2023super,kamath2023s} studying spatial reasoning focused on spatial relationships \wrt the viewer, \eg left/right relationships from the viewer's perspective. We regard these as 2D spatial relationships as they can be derived merely from 2D bounding boxes of the objects. To study how LMMs can perceive and understand 3D spatial relationships, previous datasets often adopted synthetic environments, \eg, Blender, with controllable simulation and 3D groundtruths for automatic question-answer generation~\cite{chen2022comphy,wang20243d,wang2024compositional}. However, synthetic images in these datasets exhibit a large domain gap with natural images and it remains unclear if insights and findings from these datasets would generalize to the real image domain. More recent works, such as SpatialRGPT~\cite{cheng2024spatialrgpt} and Cambrian-1~\cite{tong2024cambrian}, built on existing datasets with 3D annotations~\cite{brazil2023omni3d,song2015sunrgbd,baruch2021arkitscenes,geiger2012kitti,roberts2021hypersim,caesar2020nuscenes} and generated visual question-answer pairs with pre-defined rules. Despite the improved image quality, they are essentially limited to a small number of rigid object categories in Omni3D~\cite{brazil2023omni3d} and the automatically generated VQAs are subject to shortcuts and biases. To enable a comprehensive and robust evaluation of the 3D spatial reasoning capabilities, we manually annotate visual question-answer pairs on diverse and open-vocabulary entities, such as logos on a car or arrows on the billboard, enforcing balanced data distributions in multiple aspects and avoiding questions with shortcuts or trivial answers.

\paragraph{3D awareness of visual foundation models.} With the recent advancements in large multi-modal models~\cite{liu2023llava,liu2023improvedllava,liu2024llavanext}, there has been a rising interest in applying these LMMs to a broader range of tasks, such as chatting about human poses~\cite{feng2024chatpose}, embodied question answering~\cite{OpenEQA2023}, and robotic manipulation~\cite{huang2024rekep,huang2023voxposer}. Notably, these tasks involve reasoning and interacting with the 3D scenes, which largely builds on the 3D awareness of vision encoders. Previous works studied the 3D awareness of visual foundation models by adopting proxy tasks, such as part correspondence~\cite{el2024probing} and pose estimation~\cite{ma2024imagenet3d}, and quantitatively evaluating the 3D awareness with linear probing.
Our work can be considered as one step further --- studying the 3D recognition and reasoning capabilities of LMMs by benchmarking their performance on fundamental 3D spatial relationship questions. Future research on downstream tasks, such as automatic navigation and robotic manipulation, could refer to the findings in our \datasetname{} and adopt LMMs with better 3D spatial reasoning capabilities.

\section{\datasetname} \label{sec:dataset}

In this section we introduce \datasetname{} for comprehensively analyzing the 3D spatial reasoning capabilities of LMMs. We start by presenting the design considerations in \cref{sec:dataset-design}, \ie, how these design choices lead to a robust and valuable evaluation of 3D spatial reasoning capabilities. Then we show the four main question types in \cref{sec:dataset-types}, as well as the challenges in each type of questions. Next we introduce the three splits of \datasetname{} and their scopes in \cref{sec:dataset-splits}. In \cref{sec:dataset-eval} we present our evaluation strategies, including CircularEval and FlipEval.
Please refer to \cref{sec:dataset-stats} in supplementary materials where we provide details of our data collection and summary statistics of \datasetname.

\subsection{Design of \datasetname} \label{sec:dataset-design}

When developing \datasetname, we incorporate the following four key designs to enable a robust and valuable evaluation of 3D spatial reasoning capabilities.
\textbf{First, our 3D spatial reasoning questions are based on open-vocabulary entities.} Previous spatial reasoning benchmarks~\cite{chen2024spatialvlm,tong2024cambrian} largely relied on existing datasets with 3D annotations~\cite{brazil2023omni3d}, which limited their scope to a small number of rigid object categories. In our \datasetname, we annotate 3D spatial reasoning questions across a broad range of open-vocabulary entities (see \cref{fig:teaser}), enabling a thorough analysis of the 3D awareness and 3D reasoning capabilities of LMMs over diverse, commonly encountered real-world objects.
\textbf{Next, we avoid questions with shortcuts or trivial answers.} For instance, objects higher in 3D space are usually higher in 2D space. We collect diverse VQAs and avoid those with clear shortcuts (see \cref{fig:dataset-height}). Also, when comparing which of the two objects has a smaller 3D distance to a third anchor object, we avoid the cases when there is a significant gap between the two distances, which lead to trivial answers.
\textbf{Moreover, we implement a balanced data distribution in various aspects,} such as a roughly same number of yes/no answers and complementary image pairs~\cite{goyal2017making} that lead to opposite answers given the same 3D spatial reasoning question (see \cref{fig:teaser-b}). This effectively removes priors in the answer distribution, \eg, pedestrians are often located lower than street lights, or the fact that objects higher in 3D space are also higher in 2D image plane. This design ensures that models cannot exploit biases or shortcuts for a higher benchmark performance.
\textbf{Lastly, we adopt special evaluation strategies for robust evaluation,} including previous CircularEval~\cite{liu2025mmbench} and our novel FlipEval (see \cref{sec:dataset-eval}).

\subsection{Question Types} \label{sec:dataset-types}

We present the 4 types of 3D spatial reasoning questions in our \datasetname. We discuss why they are challenging for LMMs and what kinds of 3D awareness and 3D spatial reasoning are needed to succeed in each type of questions. We present an overview of the 4 question types in \cref{tab:dataset-types-overview}.

\begin{figure*}[t]
  \centering
  \begin{subfigure}{0.4280\linewidth}  
    \centering
    \includegraphics[width=\linewidth]{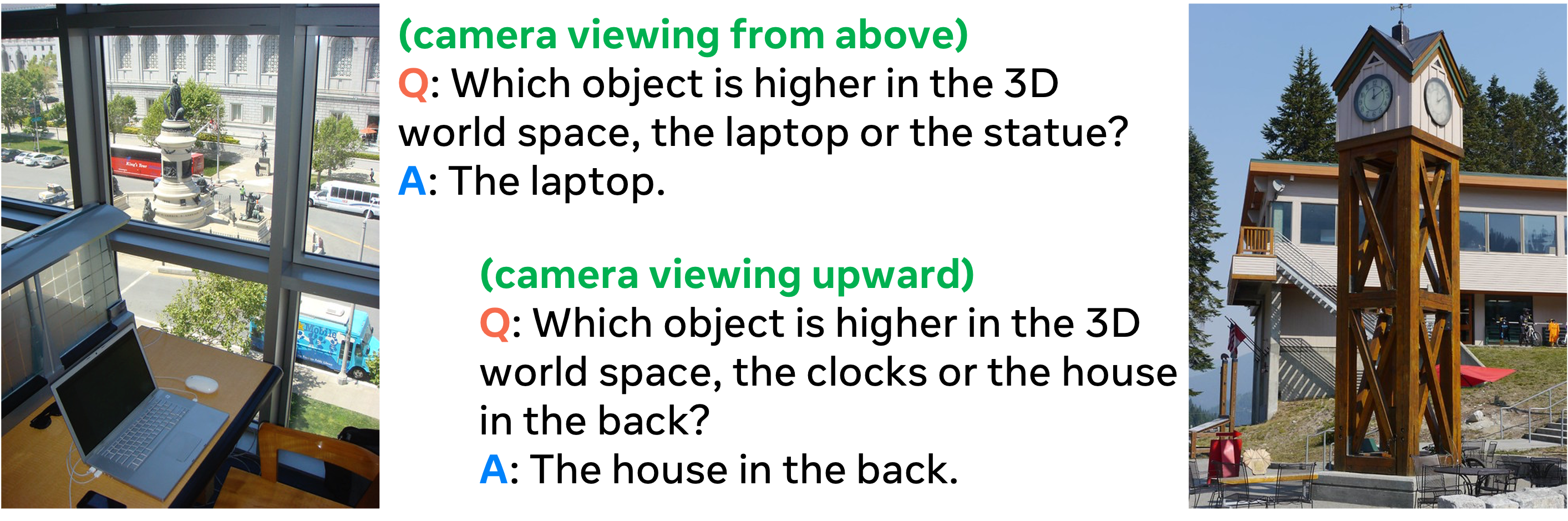}
    \caption{Height questions with different camera pitch rotations.}
    \label{fig:dataset-height}
  \end{subfigure}
  \hfill
  \begin{subfigure}{0.5030\linewidth}  
    \centering
    \includegraphics[width=\linewidth]{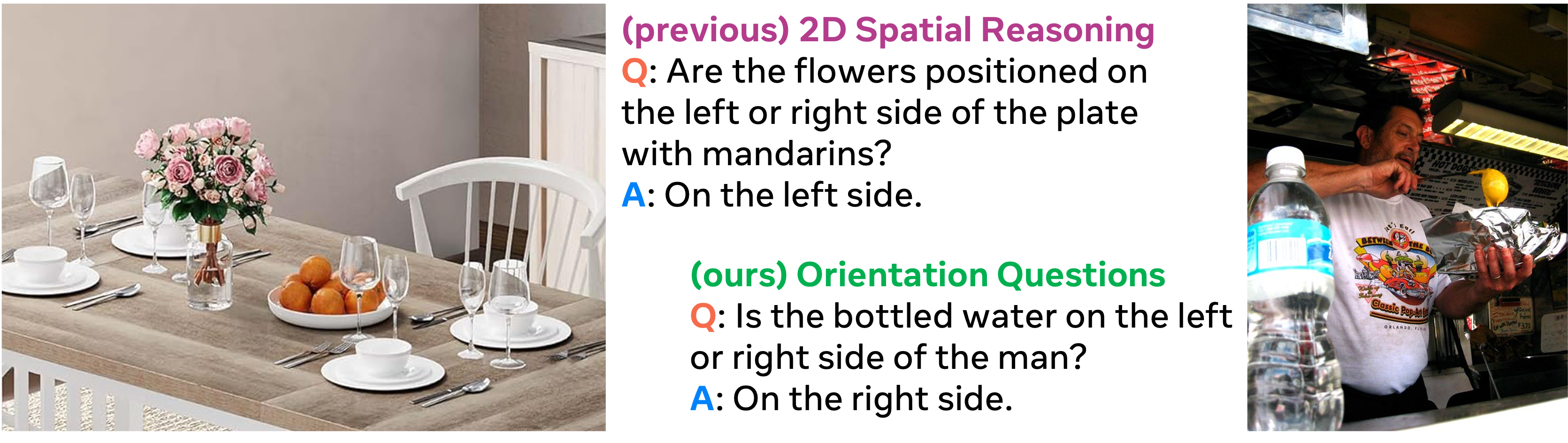}
    \caption{Comparison between 2D and 3D spatial reasoning questions.}
    \label{fig:dataset-orientation}
  \end{subfigure}
  \caption{\textbf{Challenges of 3D spatial reasoning questions in our \datasetname. See \cref{sec:dataset-types}.} \textbf{(a)} Height questions requires 3D spatial reasoning over a combination of camera extrinsics and object 3D locations. Notice how different camera pitch rotations play a crucial role to determine the final answer. \textbf{(b)} Previous 2D spatial reasoning questions can be addressed by analyzing objects' 2D locations and depths, while our orientation questions require complex 3D spatial reasoning on objects' 3D orientations and 3D locations.}
  \label{fig:challenges}
\end{figure*}

\paragraph{Height questions.} For height-related question, we study if models can determine which of the two given objects is positioned higher in the 3D world space. To correctly answer the questions, a model must (i) calibrate camera extrinsics, such as roll and pitch rotations, and then (ii) detect 3D locations of the objects in the 3D world space. This task poses a significant challenge for large multi-modal models as these fine-grained 3D knowledge are hard to derive from the weak language supervision in standard multi-modal pre-training. In Figure~\ref{fig:dataset-height} we illustrate two examples of height questions. Notice how different pitch rotations of the camera, \ie, viewing from above in the left figure and viewing upward in the right figure, play a crucial role to determine the final answer. In both examples, relying solely on the 2D locations within the image plane or the 3D locations in the camera coordinate system would lead to incorrect answers.

\paragraph{Location questions.} There are three subtypes of location-related questions, \ie, determining (i) if two objects are next to or far from each other, (ii) which of the two objects is closer to the camera, and (iii) if an object is directly above or underneath another object. Models must not only ground the 2D locations of the objects, but also understand the depth of field presented in the image. Consider the location question in \cref{fig:teaser-a}. Although the 2D locations of the man and the hydrant are close, they are in fact far away from each other in the 3D space. Humans can determine the answer by estimating a rough depths of the two objects, or from other visual cues, such as how the pedestrian walk leads towards the vanishing point. Other examples include the top two questions in \cref{fig:teaser-b}, which also require an understanding of the depth field.

\paragraph{Orientation questions.} Orientation-related questions study the 3D spatial reasoning that involves estimating the 3D orientation of an object. These questions are divided into three subtypes: determining which ``side'' of an object faces the camera, whether an object is in front of or behind another, and if an object is positioned on the left or right side of another. Unlike previous 2D spatial reasoning questions~\cite{chen2024spatialvlm} that focus on spatial relationships \wrt the viewer's perspective, our orientation-related questions emphasize spatial relationships from the object's perspective. As demonstrated in \cref{fig:dataset-orientation}, 2D spatial reasoning questions can be addressed by analyzing objects' 2D locations and depths. Meanwhile, our orientation questions require estimating objects' 3D orientation and perform 3D spatial reasoning across various dimensions of 3D information.

\paragraph{Multi-object reasoning questions.} Multi-object reasoning questions consider the 3D spatial relationships between multiple objects, such as asking which side of an object is facing another object, or with three objects, asking which of the given objects is facing towards or closer to the third object. These questions require more advanced 3D awareness than simpler 3D concepts such as ``closer'' (to the camera) or ``higher'', and require more complex 3D spatial reasoning, such as comparing distances between multiple objects from multi-step 3D computation.

\begin{table}
  \small
  \centering
  \resizebox{\columnwidth}{!}{%
  \begin{tabular}{lccccc}
    \toprule
    Type & \# Subtypes & Camera & Loc. & Orient. & Reasoning \\
    \midrule
    Height & 1 & \checkmark & \checkmark & & +\textcolor{white}{+} \\
    Location & 3 & & \checkmark & & +\textcolor{white}{+} \\
    Orientation & 3 & & \checkmark & \checkmark & +\textcolor{white}{+} \\
    Multi-Object & 5 & & \checkmark & \checkmark & ++ \\
    \bottomrule
  \end{tabular}}
  \caption{\textbf{Overview of the 4 main types of 3D spatial reasoning questions} and what kinds of 3D awareness and spatial reasoning are needed to answer each types of questions.}
  \label{tab:dataset-types-overview}
\end{table}

\subsection{Benchmark Splits} \label{sec:dataset-splits}

Our \datasetname{} is composed of three splits, a \lstinline{real} split with 2,100 3D spatial reasoning questions on MS-COCO images~\cite{lin2014microsoft} and two \lstinline{synthetic} splits with 672 questions on synthetic images rendered with 3D scenes in HSSD~\cite{khanna2024habitat}.
We evaluate the standard 3D spatial reasoning capabilities of LMMs on visual question-pairs from the \lstinline{real} split, and with the \lstinline{synthetic} split, we study the robustness of 3D spatial capabilities \wrt common and uncommon camera 6D viewpoints by analyzing the gap between the \lstinline{synthetic-common} and \lstinline{synthetic-uncommon} splits.

\begin{figure*}[t]
  \centering
  \begin{subfigure}{\linewidth}
    \centering
    \includegraphics[width=0.92\linewidth]{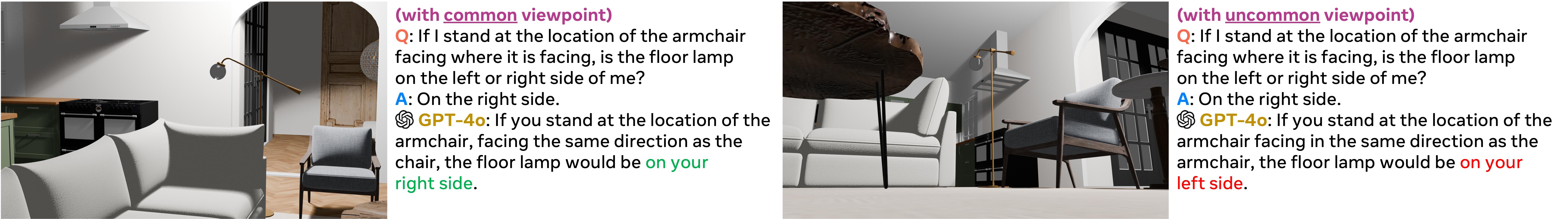}
    \caption{Orientation questions on multi-view images from common (left) and uncommon (right) camera 6D viewpoints.}
    \label{fig:hssd-a}
  \end{subfigure}
  \par\medskip
  \begin{subfigure}{\linewidth}
    \centering
    \includegraphics[width=0.92\linewidth]{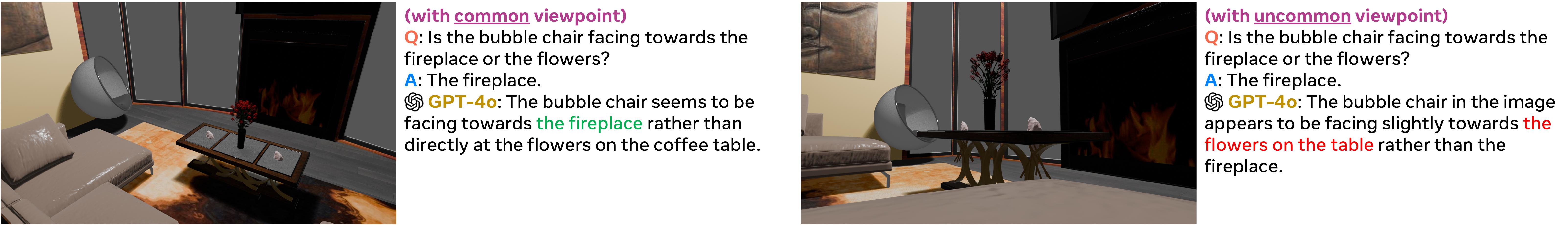}
    \caption{Multi-object reasoning questions on multi-view images from common (left) and uncommon (right) camera 6D viewpoints.}
    \label{fig:hssd-b}
  \end{subfigure}
  \caption{\textbf{Examples of the paired visual question-answer data in our \datasetname-\lstinline{synthetic}.} \textbf{(a)} Example questions from the four main types of 3D spatial reasoning questions. \textbf{(b)} To enable a robust evaluation of the 3D spatial reasoning capabilities, we collect complementary images that lead to opposite answers given the same question and adopt a novel FlipEval strategy (see \cref{sec:dataset-eval}).}
  \label{fig:hssd}
\end{figure*}

With the HSSD 3D scenes and controllable photorealistic rendering, we obtain multi-view images of the same 3D scene, each rendered with a common and an uncommon viewpoint. We ask the same 3D spatial reasoning question regarding the two images and study if models can obtain the correct answers on common and uncommon camera 6D viewpoints.
We define ``common'' viewpoints as 6D camera poses with zero roll rotation, small pitch rotation, and taken from the height of a human, simulating the typical perspective when people take pictures. Conversely, ``uncommon'' viewpoints include 6D poses with noticeable roll rotation, large pitch rotation, or perspectives taken close to the ground or from a high location.
The two synthetic splits are denoted by \lstinline{synthetic-common} and \lstinline{synthetic-uncommon} and examples from the two splits are demonstrated in \cref{fig:hssd}. Notice how the answers by GPT-4o are correct when shown the image from a common camera 6D viewpoint and wrong when prompted from an uncommon viewpoint, despite both images present a clear view of the 3D scene and humans can derive the correct answers without any difficulty.

\subsection{Evaluation} \label{sec:dataset-eval}

Since all 3D spatial reasoning questions in \datasetname{} have two or four answer choices, we formulate these questions as multiple choice questions with two or four options. To accommodate the free-form answers predicted by pretrained LMMs, we follow \cite{liu2025mmbench} and adopt LLM-involved choice extraction to obtain the predicted label. To enable a robust evaluation of various 3D spatial reasoning capabilities, we adopt the following two designs during testing:

\paragraph{CircularEval~\cite{liu2025mmbench}.} To avoid the bias of choice ordering and the influence of random guessing for multiple choice questions, we adopt CircularEval~\cite{liu2025mmbench} for more robust benchmark performance. Specifically we feed each question into the LMM two or four times, each with a different ordering of the answer choices. The LMM is considered successful in answering this question only if the predicted answer is correct for all passes.

\paragraph{FlipEval.} Following the left-right biases discussed in~\cite{Ranasinghe_2024_CVPR}, we further propose a novel \textit{FlipEval} to remove left/right biases in 3D with paired visual question-answer pairs. By applying horizontal flip to an image, we obtain a new visual question. The answer would generally remain the same, such as for location and height questions, but when it involves 3D spatial relationships such as ``left'' and ``right'', the answer would change. We illustrate this idea in \cref{fig:teaser-b}, where the elephant logo is the on the left of the truck but changes to the right after image flipping. FlipEval effectively removes left/right biases in 3D spatial relationship, such as driver often sitting on the left side of the car or most people holding tools in their right hands. Lastly FlipEval also avoids the influence of random guessing and enriches the image distribution in our \datasetname.

\section{Experiments}
\label{sec:exp}

We first introduce our experimental settings in \cref{sec:exp-settings}. Next in \cref{sec:exp-standard} we benchmark various LMMs on \datasetname{}. We further study how various model designs, \ie, choice of visual encoders and scaling of language models, attribute to the 3D spatial reasoning abilities. Then we evaluate various LMMs on our \datasetname-\lstinline{synthetic} and analyze the robustness of LMMs \wrt uncommon camera viewpoints in \cref{sec:robustness}. Lastly we present some failure cases of GPT-4o and Gemini 2.0 in \cref{sec:failure_cases}, highlighting limitations of current state-of-the-art LMMs and discussing possible future improvements.

\subsection{Experimental Settings} \label{sec:exp-settings}

With our \datasetname, we study: (1) standard 3D spatial reasoning abilities, by benchmarking various LMMs on \datasetname-\lstinline{real} with VQAs on real images from MS-COCO~\cite{lin2014microsoft}, and (2) robustness of 3D spatial reasoning abilities \wrt uncommon camera viewpoints, by analyzing the performance gap between the two \datasetname-\lstinline{synthetic} splits with common and uncommon viewpoints.


\paragraph{Testing data augmentation.} We develop rule-based methods to augment the annotated visual question-answer pairs and obtain a larger number of testing data with a balanced and rich set of 3D spatial relationships. For instance, given a question asking which object is higher in the 3D world space, we generate a new question asking which object is lower in the 3D world space. We further adopt FlipEval that augments the question set by horizontally flipping the images. This leads to a total of 5,250 questions on MS-COCO images, \ie, \datasetname-\lstinline{real}, and 1,692 questions on synthetic images, \ie, \datasetname-\lstinline{synthetic}.

\paragraph{Evaluation.} To evaluate the correctness of free-form answers, we follow MMBench~\cite{liu2025mmbench} and use exact matching to parse choice labels, or LLM-assisted evaluation, \eg, with \lstinline{gpt-4}, when matching fails. We further adopt CircularEval~\cite{liu2025mmbench} that repeats a question $N$ times, each with a different ordering of the choices. $N$ is the number of choices.

\subsection{Results on 3D Spatial Reasoning Abilities} \label{sec:exp-standard}

\begin{table}
  \small
  \centering
  \resizebox{\columnwidth}{!}{%
  \begin{tabular}{lcccccc}
    \toprule
    \multirow{2.5}{*}{{Model}} & & \multicolumn{5}{c}{\datasetname-\lstinline{real}} \\
    \cmidrule{3-7}
    & & Overall & Height & Loc. & Orient. & Multi. \\
    \midrule
    \textbf{\textit{Baselines}} \\
    Random & & 20.9 & 25.0 & 25.0 & 16.8 & 20.1 \\
    Random++ & & 45.8 & 50.0 & 50.0 & 41.7 & 45.0 \\
    Human & & 95.7 & 92.9 & 96.4 & 97.7 & 94.9 \\
    \midrule
    \textbf{\textit{Open-sourced}} \\
    LLaVA-v1.5-7B~\cite{liu2023improvedllava} & 13 & 38.1 & 39.1 & 46.9 & 28.7 & 34.7 \\
    Cambrian-1-8B~\cite{tong2024cambrian} & 11 & 42.2 & 23.2 & 53.9 & 35.9 & 41.9 \\
    LLaVA-NeXT-8B~\cite{liu2024llavanext} & 6 & 48.4 & 50.6 & 59.9 & 36.1 & 43.4 \\
    InternVL2.5-8B~\cite{chen2024expanding} & \cellcolor[rgb]{0.780,0.892,0.997}{4} & 50.9 & 45.9 & 68.1 & 38.7 & 43.3 \\
    QWen2.5-VL-7B~\cite{Qwen2.5-VL} & 6 & 48.4 & 44.1 & 62.7 & 40.6 & 40.5 \\
    \midrule
    \textbf{\textit{Specialist}} \\
    SpatialLLM~\cite{ma2025spatialllm} & 9 & 44.8 & 45.8 & 61.6 & 30.0 & 36.7 \\
    SpatialRGPT~\cite{cheng2024spatialrgpt} & 14 & 32.7 & 55.9 & 39.0 & 27.8 & 20.0 \\
    SpatialRGPT \textit{w/} depth~\cite{cheng2024spatialrgpt} & 6 & 48.4 & 55.9 & 60.0 & 34.2 & 42.3 \\
    SpatialReasoner~\cite{ma2025spatialreasoner} & \cellcolor[rgb]{0.443,0.727,0.991}{1} & 60.3 & 52.5 & 75.2 & 55.2 & 51.8 \\
    \midrule
    \textbf{\textit{Proprietary}} \\
    Claude-3.5V-Sonnet~\cite{claude} & 7 & 48.2 & 53.5 & 63.1 & 31.4 & 41.3 \\
    Gemini-2.0-Flash~\cite{google2024gemini2} & \cellcolor[rgb]{0.890,0.946,0.998}{5} & 49.8 & 49.7 & 68.9 & 32.2 & 41.5 \\
    Gemini-2.0-Flash-bbox~\cite{google2024gemini2} & 8 & 47.5 & 45.2 & 66.5 & 27.7 & 41.4 \\
    Gemini-2.0-Flash-think~\cite{google2024gemini2} & \cellcolor[rgb]{0.667,0.837,0.995}{3} & 51.1 & 53.0 & 67.1 & 35.8 & 43.6 \\
    GPT-4o-mini~\cite{hurst2024gpt4o} & 12 & 39.7 & 44.3 & 52.4 & 21.0 & 36.5\\
    GPT-4o~\cite{hurst2024gpt4o} & 10 & 44.2 & 53.2 & 59.6 & 21.6 & 39.0 \\
    QWenVLMax~\cite{Qwen2.5-VL} & \cellcolor[rgb]{0.557,0.783,0.993}{2} & 52.0 & 45.1 & 70.7 & 37.7 & 44.8 \\
    \bottomrule
  \end{tabular}}
  \caption{\textbf{Experimental comparison of state-of-the-art large multi-modal models on our \datasetname.} Results show that state-of-the-art LMMs exhibit limited 3D spatial reasoning capabilities. Please refer to \cref{sec:exp-standard} for detailed analyses.}
  \label{tab:main_results}
\end{table}

\begin{table*}
  \small
  \centering
  \resizebox{0.9\textwidth}{!}{%
  \begin{tabular}{lllccccccccc}
    \toprule
    & & & & \multicolumn{5}{c}{\datasetname} \\
    \cmidrule{5-9}
    LLM & Vision Encoder & Connector & & Mean & Height & Loc. & Orient. & Multi. \\
    \midrule
    \textbf{\textit{Baseline}} \\
    Vicuna-v1.5-7B~\cite{zheng2023judging, liu2023improvedllava} & CLIP-L14-336~\cite{radford2021learning} & 2xMLP & & 36.8 & 38.5 & \textbf{46.4} & 27.7 & 31.8 \\
    \midrule
    \multicolumn{2}{l}{\textbf{\textit{Mixed Encoders}}} \\
    Vicuna-v1.5-7B~\cite{zheng2023judging, liu2023improvedllava} & CLIP-L14-336~\cite{radford2021learning} + DINOv2-L14-224~\cite{oquab2023dinov2} & 2xMLP & & \underline{37.2} & \underline{45.9} & 42.2 & \textbf{28.7} & \underline{33.6} \\
    Vicuna-v1.5-7B~\cite{zheng2023judging, liu2023improvedllava} & CLIP-L14-336~\cite{radford2021learning} + MAE-H14~\cite{he2022masked} & 2xMLP & & 33.1 & 42.7 & 39.2 & 26.1 & 27.5 \\
    Vicuna-v1.5-7B~\cite{zheng2023judging, liu2023improvedllava} & CLIP-L14-336~\cite{radford2021learning} + SAM-L~\cite{kirillov2023segment} & 2xMLP & & 27.9 & 44.6 & 34.4 & 16.5 & 21.5 \\
    \midrule
    \multicolumn{2}{l}{\textbf{\textit{Connectors}}} \\
    Vicuna-v1.5-7B~\cite{zheng2023judging, liu2023improvedllava} & CLIP-L14-336~\cite{radford2021learning} + DINOv2-L14-224~\cite{oquab2023dinov2} & SVA~\cite{tong2024cambrian} & & \textbf{37.8} & \textbf{46.0} & \underline{43.1} & 26.5 & \textbf{35.9} \\
    Vicuna-v1.5-7B~\cite{zheng2023judging, liu2023improvedllava} & CLIP-L14-336~\cite{radford2021learning} + MAE-H14~\cite{he2022masked} & SVA~\cite{tong2024cambrian} & & 34.1 & 45.3 & 38.6 & 25.3 & 30.2 \\
    \bottomrule
  \end{tabular}}
  \caption{\textbf{Experimental results on LMMs with various vision encoder setups.} We use \lstinline{LLaVA-v1.5-7B} as the baseline model and studies how vision encoders with different features contribute to the final 3D spatial reasoning abilities of LMMs.
  }
  \label{tab:variants}
\end{table*}

\begin{table*}
  \small
  \centering
  \resizebox{0.95\textwidth}{!}{%
  \begin{tabular}{lccccccccccccc}
    \toprule
    \multirow{2.5}{*}{{Model}} & \multicolumn{5}{c}{\datasetname{}-\lstinline{synthetic-common}} & & \multicolumn{5}{c}{\datasetname{}-\lstinline{synthetic-uncommon}} & & Rel. Drop \\
    \cmidrule{2-6} \cmidrule{8-12} \cmidrule{14-14}
    & Overall & Height & Loc. & Orient. & Multi. & & Overall & Height & Loc. & Orient. & Multi. & & $\delta$ \\
    \midrule
    \textbf{\textit{Open-sourced}} \\
    LLaVA-v1.5-7B~\cite{liu2023llava} & 42.0 & 40.0 & 50.6 & 20.8 & 47.6 & & 38.0 & 41.0 & 43.6 & 17.9 & 45.2 & & \textcolor{myred}{-9.5\%} \\
    Cambrian-1-8B~\cite{tong2024cambrian} & 48.1 & 37.5 & 56.1 & \underline{39.6} & 47.6 & & 39.9 & 35.0 & 45.7 & 29.2 & 41.9 & & \textcolor{myred}{-17.0\%} \\
    LLaVA-NeXT-8B~\cite{liu2024llavanext} & 45.5 & \underline{65.0} & 57.9 & 10.4 & 50.0 & & 36.8 & 47.5 & 44.5 & 7.3 & 46.0 & & \textcolor{myred}{-19.1\%} \\
    \midrule
    \textbf{\textit{Proprietary}} \\
    Qwen-VL-Plus~\cite{bai2023qwen} & 30.7 & 35.0 & 37.8 & 30.2 & 20.2 & & 21.0 & 15.0 & 25.0 & 22.9 & 16.1 & & \textcolor{myred}{-31.6\%} \\
    Qwen-VL-Max~\cite{bai2023qwen} & \underline{55.2} & 62.5 & \underline{69.5} & 31.2 & \underline{52.4} & & \underline{48.6} & 52.5 & \textbf{59.8} & 24.0 & \underline{51.6} & & \textcolor{myred}{-12.0\%} \\
    Claude-Sonnet~\cite{claude3} & 47.4 & 47.5 & 58.5 & 26.0 & 49.2 & & 39.4 & \textbf{60.0} & 48.2 & 16.7 & 38.7 & & \textcolor{myred}{-16.9\%} \\
    Gemini-1.5-Flash~\cite{team2024gemini1-5} & 44.6 & 57.5 & 59.8 & 13.5 & 44.4 & & 37.7 & 42.5 & 45.7 & 11.5 & 46.0 & & \textcolor{myred}{-15.6\%} \\
    Gemini-1.5-Pro~\cite{team2024gemini1-5} & \textbf{59.9} & \underline{65.0} & 69.5 & \textbf{50.0} & \textbf{53.2} & & \textbf{49.5} & 42.5 & 52.4 & \textbf{40.6} & \textbf{54.8} & & \textcolor{myred}{-32.2\%} \\
    GPT-4o-mini~\cite{hurst2024gpt4o} & 46.5 & 47.5 & 53.7 & 36.5 & 44.4 & & 40.3 & 42.5 & 43.9 & \underline{33.3} & 40.3 & & \textcolor{myred}{-13.3\%} \\
    GPT-4o~\cite{hurst2024gpt4o} & 51.2 & \textbf{70.0} & \textbf{70.1} & 17.7 & 46.0 & & 44.3 & \textbf{60.0} & \underline{58.5} & 15.6 & 42.7 & & \textcolor{myred}{-13.5\%} \\
    \bottomrule
  \end{tabular}}
  \caption{\textbf{Experimental results on our \datasetname-\lstinline{synthetic-common} and \datasetname-\lstinline{synthetic-uncommon}.} We study the robustness of 3D spatial reasoning capabilities of LMMs by analyzing the performance gap between the two splits with images from the same 3D scene but from ``common'' and ``uncommon'' viewpoints. We find that LMMs does not generalize well to images with 6D camera viewpoints less represented in their training set. See \cref{sec:robustness} for detailed discussions.}
  \label{tab:results-robustness}
\end{table*}

We benchmark a wide range of open-sourced and proprietary LMMs on our \datasetname-\lstinline{real} and analyze 3D spatial reasoning abilities on different types of questions. We consider three baseline results:
(i) \textbf{random}: a simple baseline that predicts random answers for all visual questions.
(ii) \textbf{random++}: a stronger random baseline that predicts consistent answers given different choice orders of a same visual question in CircularEval.
(iii) \textbf{human}: a human-level performance established by human evaluators that did not participate in the data annotation process.
We report the full results in \cref{tab:main_results}.

We make the following observations: (i) \textbf{State-of-the-art LMMs have limited 3D spatial reasoning capabilities}, as found by low performance achieved by state-of-the-art open-sourced and proprietary LMMs, falling far behind human-level performance. (ii) \textbf{Scaling laws for LMMs are not effective for 3D spatial reasoning.} Results show that despite significant more training data and computation spent on the proprietary LMMs, they demonstrate limited advantages over open-sourced counterparts, featuring high-quality data with efficient training setups. Standard scaling laws demonstrate diminishing returns for 3D spatial reasoning abilities and we believe more effective approaches, \eg, 3D-aware data, architecture, and training, would be necessary to significantly advance 3D spatial reasoning.

\paragraph{Design choices of visual encoder.} We study how design choices of visual encoders can benefit 3D spatial reasoning abilities. Built on \lstinline{LLaVA-v1.5-7B}~\cite{liu2023llava}, we experiment on a range of models with different choices of visual foundation models, \ie, CLIP~\cite{radford2021learning}, MAE~\cite{he2022masked}, DINOv2~\cite{oquab2023dinov2}, SAM~\cite{kirillov2023segment}, or model designs, \ie, mixed encoders and visual projectors. Results in \cref{tab:variants} show that with mixed encoders, DINOv2 can improve the overall 3D spatial reasoning abilities of LMMs, specifically for orientation and multi-object reasoning questions that build heavily on object 3D orientations.
We also notice significant improvements for height questions when adopting MAE and SAM as vision encoder, suggesting that having richer visual features could help localize objects better. With spatial vision aggregator (SVA)~\cite{tong2024cambrian}, we can further improve the LMM with mixed encoder from 37.2\% to 37.8\%, demonstrating that fusing the semantic features with 3D-aware features from DINOv2 would benefit subsequent reasoning.



\begin{figure}
    \centering
    \includegraphics[width=0.9\columnwidth]{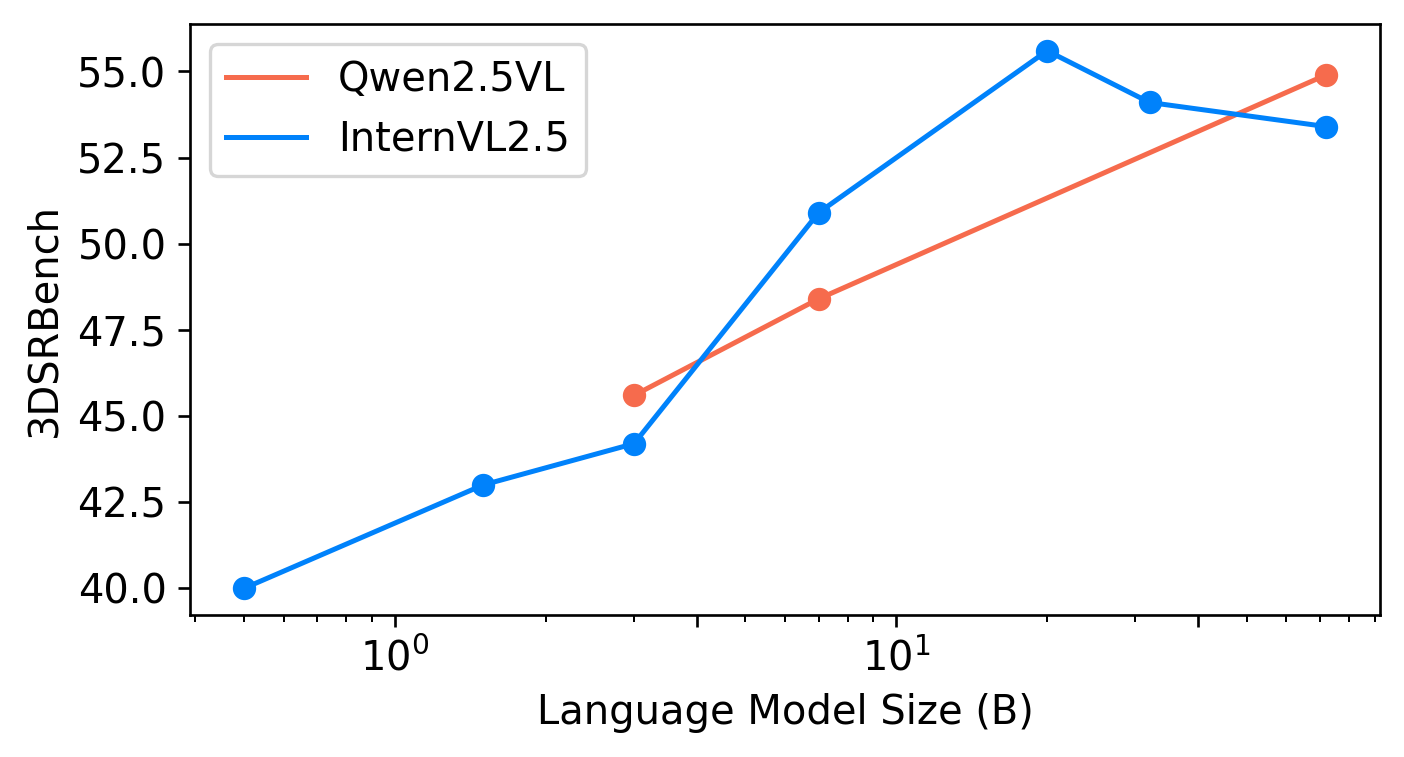}
    \caption{\textbf{Scaling of language model sizes.} Results show that scaling language model sizes can effectively improve the 3D spatial reasoning abilities of LMMs. However, with a 72B language model and a 6B vision encoder, InternVL2.5 still falls far behind human-level performance by more than 40\%.}
    \label{fig:llm-scaling}
\end{figure}

\paragraph{Scaling of language model size.} We study how the scaling of language model, \ie, in terms of the number of parameters, helps improve the 3D spatial reasoning abilities of LMMs. We consider two series of open-sourced LMMs, QWen2.5~\cite{Qwen2.5-VL} and InternVL2.5~\cite{chen2024expanding}, with a range of language model sizes from 0.5B to 72B. From the results in \cref{fig:llm-scaling}, we see that the scaling of language model sizes effectively improves the 3D spatial reasoning abilities of LMMs. Larger language models with more parameters exhibit enhanced reasoning abilities. They better capture 3D-aware information from the visual features and perform more complicated 3D spatial reasoning. However, given the importance of 3D spatial reasoning in a broad range of applications, scaling up language model size is highly inefficient --- LMMs with over 70B parameters exceed the computation capacity of common robotics or embodied AI systems and significantly limit the model throughput.

\subsection{Robustness to Uncommon Camera Viewpoints} \label{sec:robustness}

\begin{figure*}[t]
  \centering
  \begin{subfigure}{\linewidth}
    \centering
    \includegraphics[width=0.82\linewidth]{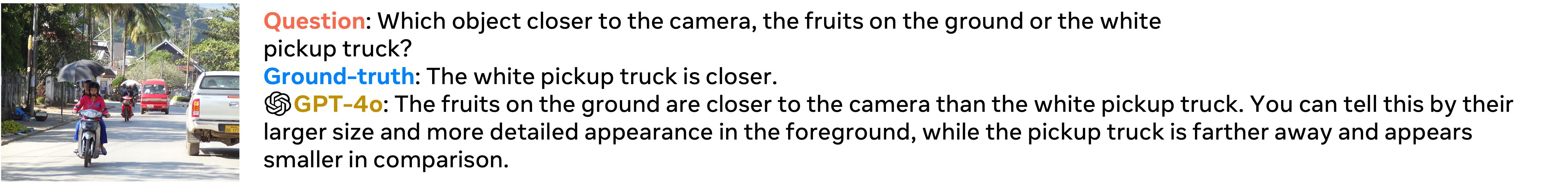}
    \caption{Failure case of GPT-4o~\cite{hurst2024gpt4o}.}
    \label{fig:failure_cases-a}
  \end{subfigure}
  \par\medskip
  \begin{subfigure}{\linewidth}
    \centering
    \includegraphics[width=0.82\linewidth]{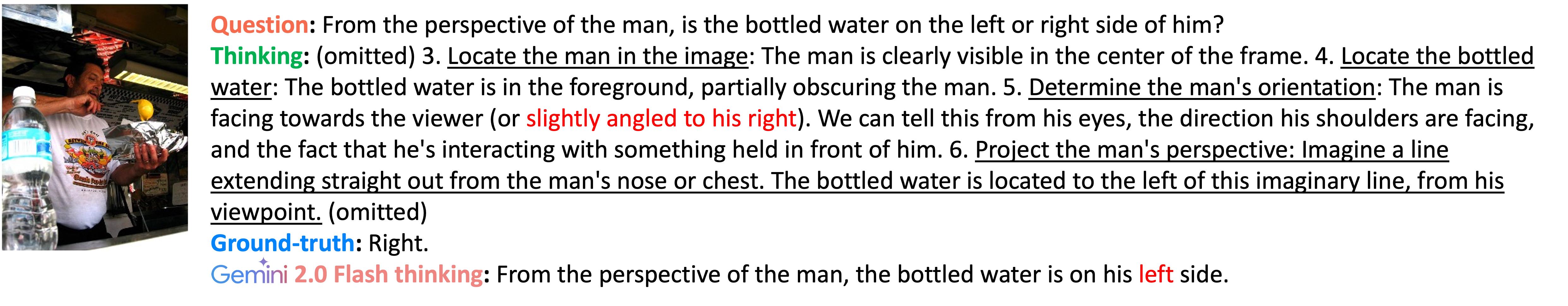}
    \caption{Failure case of Gemini 2.0 Flash thinking~\cite{google2024gemini2}.}
    \label{fig:failure_cases-b}
  \end{subfigure}
  \caption{\textbf{Failures cases of GPT-4o~\cite{hurst2024gpt4o} (top) and Gemini 2.0 Flash thinking~\cite{google2024gemini2} (bottom) on our \datasetname.} \textbf{(a) GPT-4o:} GPT-4o does not have an explicit 3D representation, \eg, metric depth, and resort to visual cues to compare the distance, which leads to a wrong answer.  \textbf{(b) Gemini 2.0 Flash thinking:} In this example Gemini 2.0 Flash thinking successfully breaks down the 3D spatial reasoning question into small and tractable steps. However, without explicit 3D representations, the model cannot perform reliable 3D spatial reasoning and predicts a wrong answer. See \cref{sec:supp-failure-cases} in supplementary materials for more failure cases of the two models.}
  \label{fig:failure_cases}
\end{figure*}

We study the robustness of 3D spatial reasoning abilities \wrt common and uncommon viewpoints. We evaluate a variety of open-sourced and proprietary LMMs on our \datasetname-\lstinline{synthetic-common} and \datasetname-\lstinline{synthetic-uncommon} splits and analyze the relative performance drop, given by
\begin{align*}
    \delta = \frac{\text{Acc}_\text{uncommon} - \text{Acc}_\text{common}}{\text{Acc}_\text{common}}
\end{align*}
As demonstrated by the results in \cref{tab:results-robustness}, all LMMs exhibit significantly degraded performance when generalizing from common to uncommon viewpoints, \eg, a 13.5\% drop in accuracy for GPT-4o~\cite{hurst2024gpt4o}, a 32.2\% drop for Gemini-1.5-Pro~\cite{team2024gemini1-5}, and a 19.1\% drop for LLaVA-NeXT-8B~\cite{liu2024llavanext}. We visualize two failure examples of GPT-4o in \cref{fig:hssd}, showing how it is capable of predicting the correct answer when prompting with an image from a common viewpoint but fails when asked the same question with an image rendered from an uncommon viewpoint of the exact same scene.

We attribute such degraded performance in uncommon viewpoints to two factors: (i) \textit{image domain gap due to different camera viewpoint distributions} between LMM training data and our \datasetname-\lstinline{synthetic-uncommon}, and (ii) \textit{state-of-the-art LMMs adopt an implicit representation of 3D scenes}.
They are heavily built on the scaling law of data-driven approaches and lack explicit 3D representations that enable reliable 3D spatial reasoning. Despite the success of data-driven methods on a range of academic and empirical benchmarks, they face severe challenges generalizing to less represented data, which in our case, are images from uncommon camera 6D viewpoints.


These findings show that 3D spatial reasoning abilities of state-of-the-art LMMs are not robust to uncommon camera viewpoints. \textbf{This largely limits their applicability to various downstream applications in robotics and embodied AI.} Cameras mounted on robot arms or embodied AI systems are often positioned in uncommon locations and orientations as used in our study (see \cref{fig:hssd}). On the one hand impressive advancements achieved by state-of-the-art LMMs in standard spatial reasoning benchmarks \cite{johnson2017clevr,hudson2019gqa,li2023super} may not generalize to downstream tasks; on the other hand, significantly degraded performance in uncommon viewpoints raises serious concerns about AI safety~\cite{amodei2016concrete}.

\subsection{Failure Cases} \label{sec:failure_cases}

We present two failure cases of GPT-4o~\cite{hurst2024gpt4o} and Gemini 2.0 Flash thinking~\cite{google2024gemini2} in \cref{fig:failure_cases}.
In \cref{fig:failure_cases-a} we see that GPT-4o cannot perform rigorous 3D spatial reasoning and resort to various visual cues for reasoning. This is because GPT-4o lacks explicit 3D representations, \eg, metric depth, that limits its ability to perform complex 3D spatial reasoning.
In \cref{fig:failure_cases-b}, Gemini 2.0 Flash thinking successfully breaks down the 3D reasoning question into small and tractable steps. However, without explicit 3D representations, the model cannot perform reliable 3D spatial reasoning step-by-step. Despite the good thinking, the model fails to follow the planning and predicts a wrong answer.

We argue that for 3D spatial reasoning problems, models must not only have strong visual encoders to parse 3D-aware features, but also build a powerful reasoning model on various 3D information. Although scaling language model size leads to stronger reasoning abilities (see \cref{fig:llm-scaling}), a lack of explicit 3D representations would fundamentally limit models' abilities to solve complex 3D spatial reasoning questions that require multi-step 3D computations.


\section{Conclusions} \label{sec:conclusions}

In this work we study the 3D spatial reasoning capabilities of LMMs. We introduce a new benchmark, \datasetname, by manually annotating 2,100 visual question-answer pairs on natural images from MS-COCO, featuring diverse and open-vocabulary entities and a balanced data distribution for robust evaluation.
To study the robustness of 3D spatial reasoning capabilities \wrt camera 6D viewpoints, we further annotate 672 visual question-answer pairs on synthetic multi-view images, each with a common and an uncommon camera viewpoint.
We benchmark a wide variety of open-sourced and proprietary LMMs on our \datasetname, studying various 3D spatial reasoning capabilities, \eg, height, location, orientation, and multi-object reasoning, as well as the robustness of these LMMs to uncommon camera viewpoints. We also study how various designs of visual encoders and scaling of language models benefit 3D spatial reasoning. Experimental results on \datasetname{} provide valuable findings and insights to develop LMMs with strong 3D spatial reasoning abilities, as well as selecting LMMs for downstream applications that require robust 3D spatial reasoning.

\section*{Acknowledgements}

We would like to thank Yiyan Li, Lizhi Ma, and the anonymous reviewers for their helpful comments and suggestions.
Wufei Ma and Alan Yuille acknowledges support from ONR with N00014-23-1-2641 and ARL award with W911NF2320008.

{
    \small
    \bibliographystyle{ieeenat_fullname}
    \bibliography{main}
}
\clearpage
\setcounter{page}{1}
\setcounter{section}{0}
\renewcommand{\thesection}{\Alph{section}}

\section{\datasetname{} Data Card} \label{sec:dataset-stats}

We employ ten annotators to annotate a total of 2,772 unique visual question-answer pairs across 12 question types. We follow the annotation principles as discussed in \cref{sec:dataset-design} and adopt an two-stage pipeline to ensure various criteria are met. Specifically after annotations are collected in the first stage, we review the quality of the collected data and reject samples with low quality or ones that lead to imbalanced data distribution. Additional new annotations are collected if necessary. Furthermore, we collect human responses for all visual question-answer pairs and disregard samples that don't reach consensus by human annotators.

\paragraph{Dataset statistics.} We annotate a total of 2,100 questions on natural images from MS-COCO~\cite{lin2014microsoft} and 674 questions on synthetic images rendered from HSSD dataset~\cite{khanna2024habitat}. With testing data augmentation and FlipEval~\cref{sec:exp-settings}, we obtain a total of 5,250 questions in \datasetname-\lstinline{real} and 1,692 questions in \datasetname-\lstinline{synthetic}. These questions are further evaluated with CircularEval~\cite{liu2025mmbench}.

\paragraph{Ethics.} We follow the ethics guidelines and obtained Institutional Review Board (IRB) approvals prior to the start of our work. We described potential risks to the annotators, such as being exposed to inappropriate images from the MS-COCO dataset, and explained the purpose of the study and how the collected data will be used. All annotators are paid by a fair amount as required at our institution.

\paragraph{License.} Our dataset is released under the CC-BY-4.0 license.

\section{Baseline Models}

\paragraph{Proprietary LMMs.} To analyze the 3D spatial reasoning capabilities of state-of-the-art LMMs, we explore a variety of proprietary LMMs, \eg, QWen-VL~\cite{Qwen2.5-VL}, Claude~\cite{claude}, Gemini~\cite{google2024gemini2}, and GPT-4o~\cite{hurst2024gpt4o}. For Gemini-2.0-Flash, we further evaluate two variants: (1) for Gemini-2.0-Flash-bbox we first ask the model to detect the 3D bounding box and then prompt the model to answer the question based on the 3D bounding boxes, and (2) for Gemini-2.0-Flash-think we evaluate the thinking model where the model performs deep thinking prior to answering the question.

\paragraph{LLaVA-v1.5~\cite{liu2023improvedllava}.} LLaVA-v1.5-7B is a strong open-sourced LMM baseline built on a Vicuna-v1.5 LLM and a CLIP-ViT-L vision encoder. It extends the visual instruction tuning framework~\cite{liu2023llava} with an MLP connector and a scaled up image resolution.

\paragraph{Cambrian-1~\cite{tong2024cambrian}.} Cambrian-1 is a strong LMM with vision-centric designs. It features an advanced connector design, spatial vision aggregator (SVA), and high-quality visual-instruction tuning data.

\paragraph{LMMs with various vision encoder designs.} We further experiment on a family of LMMs, extending the LLaVA-v1.5 baseline with various visual encoders design. We study the impact of the 3D awareness of visual encoders on the final 3D spatial reasoning capabilities. We adopt the same LLM and training strategy, exploring: (i) different mixed encoders: involving a second visual encoder besides CLIP, \eg, DINOv2~\cite{oquab2023dinov2}, MAE~\cite{he2022masked}, and SAM~\cite{kirillov2023segment}; and (ii) different visual connectors: a standard MLP connector and spatial vision aggregator (SVA)~\cite{tong2024cambrian}.

\section{Qualitative Examples of \datasetname}

\begin{figure*}[t]
    \centering
    \includegraphics[width=\textwidth]{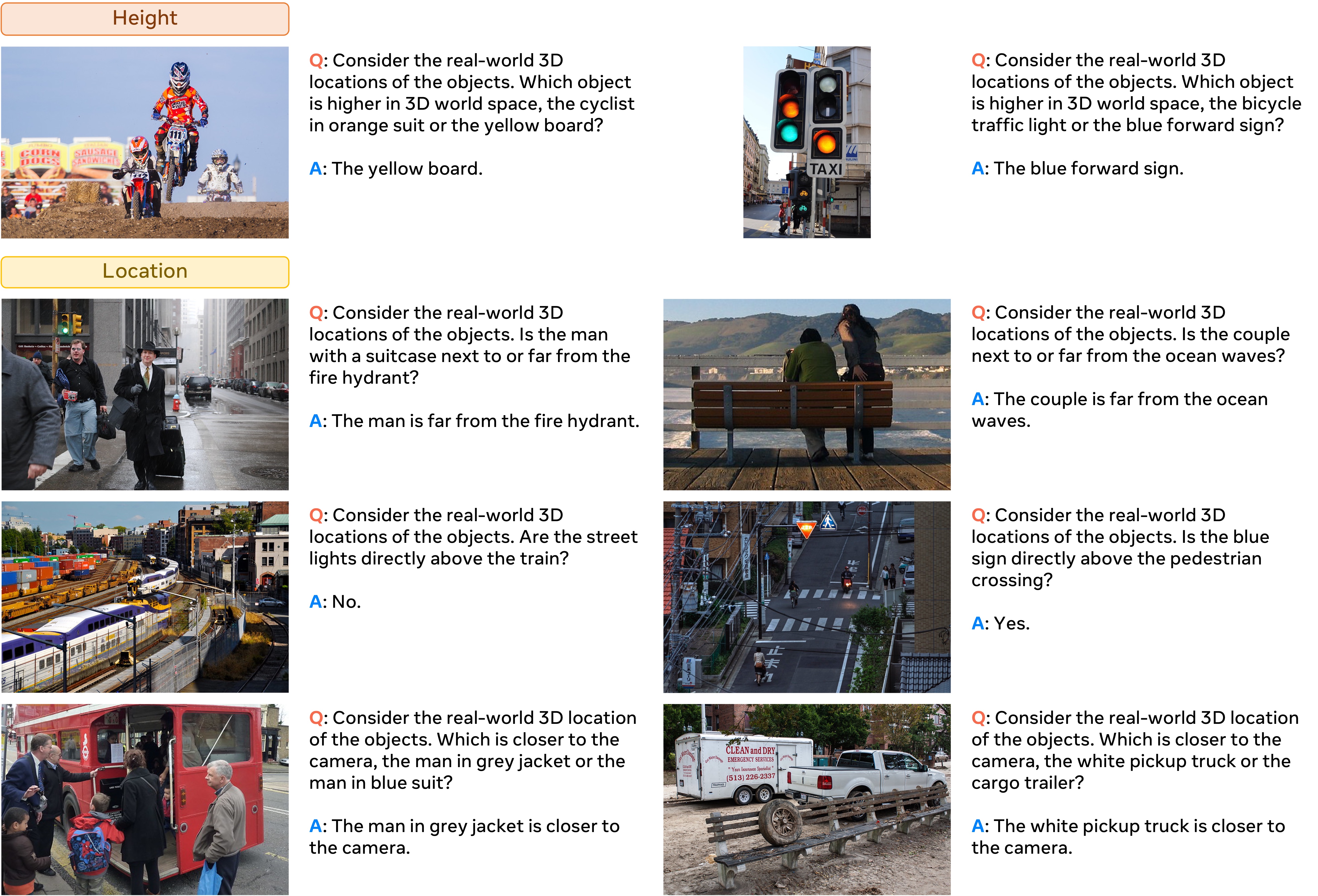}
    \caption{\textbf{Two example questions for each of the 12 question types (part I):} height and location questions.}
    \label{fig:egs-01}
\end{figure*}

\begin{figure*}[t]
    \centering
    \includegraphics[width=\textwidth]{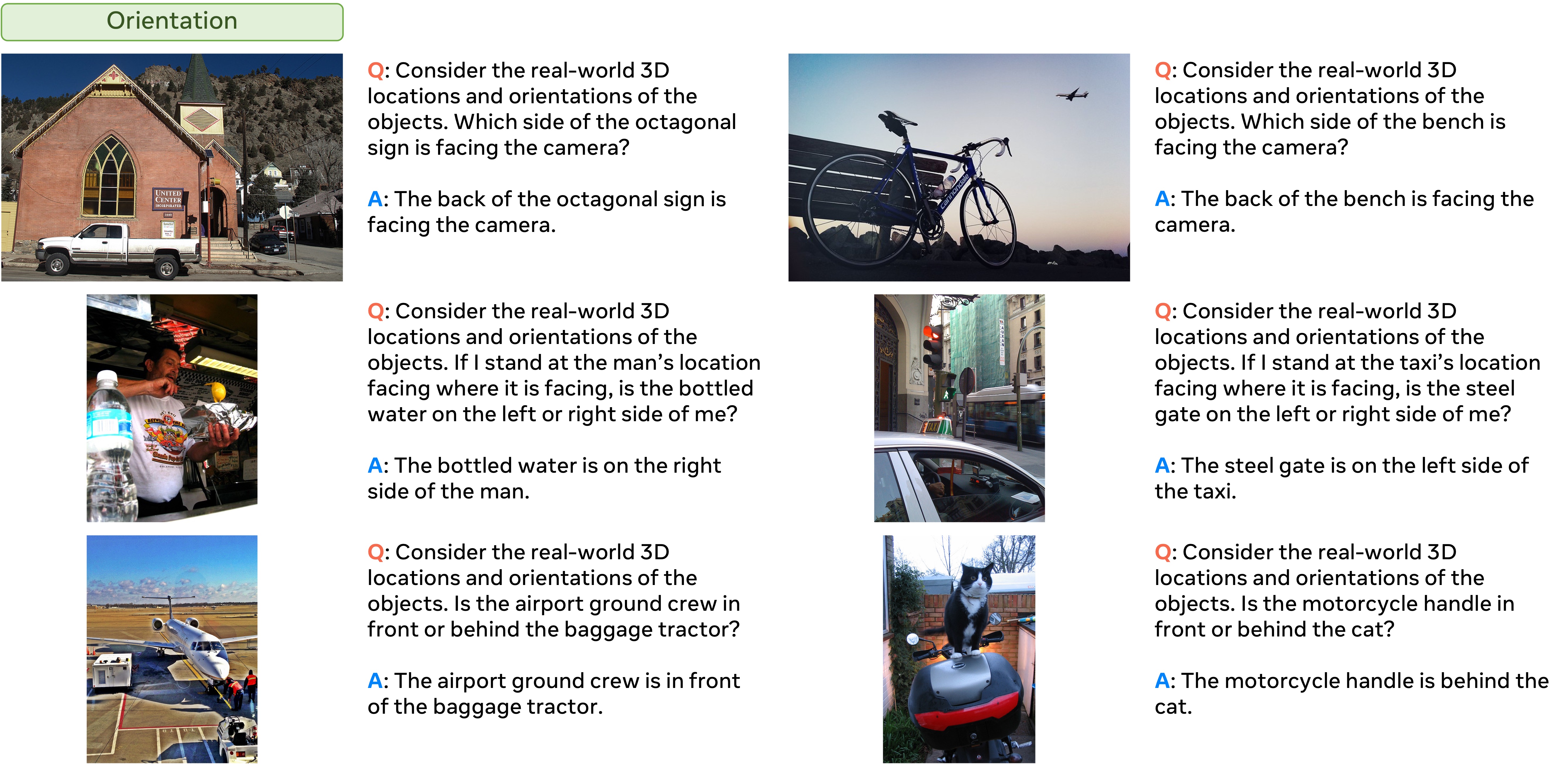}
    \caption{\textbf{Two example questions for each of the 12 question types (part II):} orientation questions.}
    \label{fig:egs-02}
\end{figure*}

\begin{figure*}[t]
    \centering
    \includegraphics[width=\textwidth]{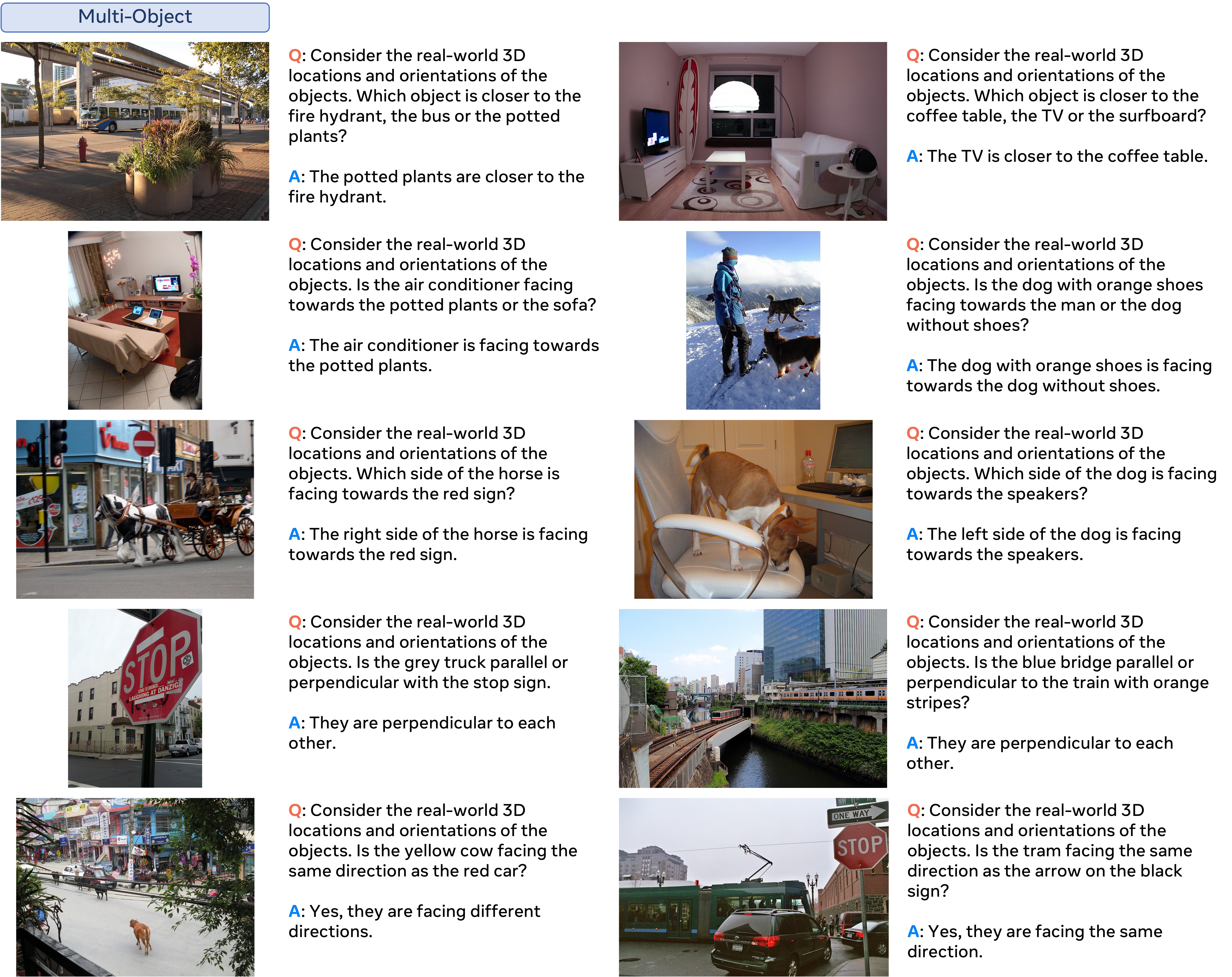}
    \caption{\textbf{Two example questions for each of the 12 question types (part III):} multi-object reasoning questions.}
    \label{fig:egs-03}
\end{figure*}

We present two example questions for each of the 12 question types in \cref{fig:egs-01} (height and location questions), \cref{fig:egs-02} (orientation questions), and \cref{fig:egs-03} (multi-object reasoning questions).

\section{Qualitative Examples of Common and Uncommon Viewpoints}

\begin{figure*}[t]
    \centering
    \includegraphics[width=0.75\textwidth]{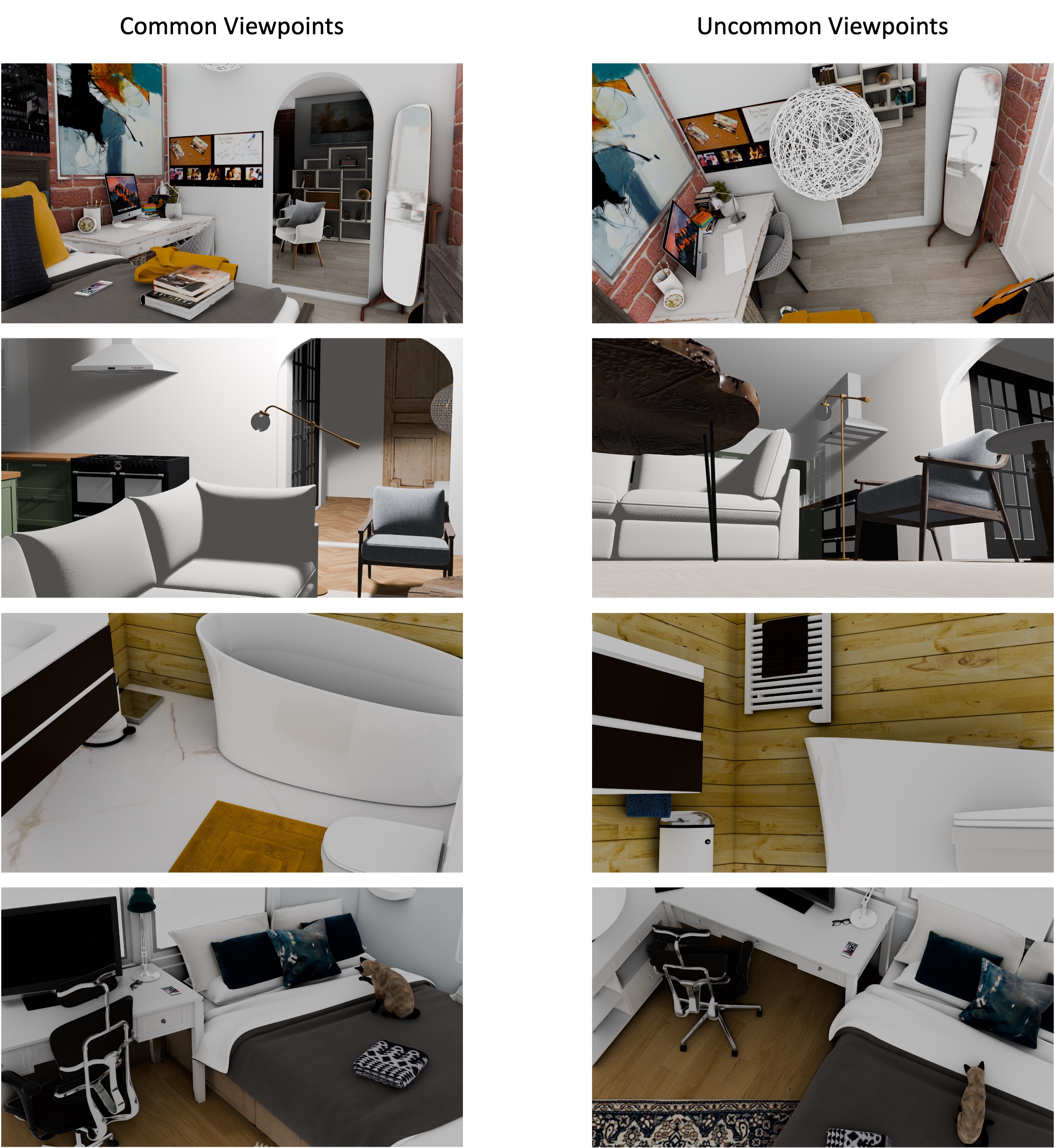}
    \caption{\textbf{Qualitative examples of multi-view images rendering the same scene and objects but from common (left) and uncommon (right) viewpoints.}}
    \label{fig:qualitative_viewpoints}
\end{figure*}

We present some qualitative examples of \datasetname-\lstinline{synthetic} with multi-view images rendering the same scene and objects but from common (left) and uncommon (right) viewpoints in \cref{fig:qualitative_viewpoints}.

\section{Failure Cases} \label{sec:supp-failure-cases}

\begin{figure*}
    \centering
    \includegraphics[width=0.9\textwidth]{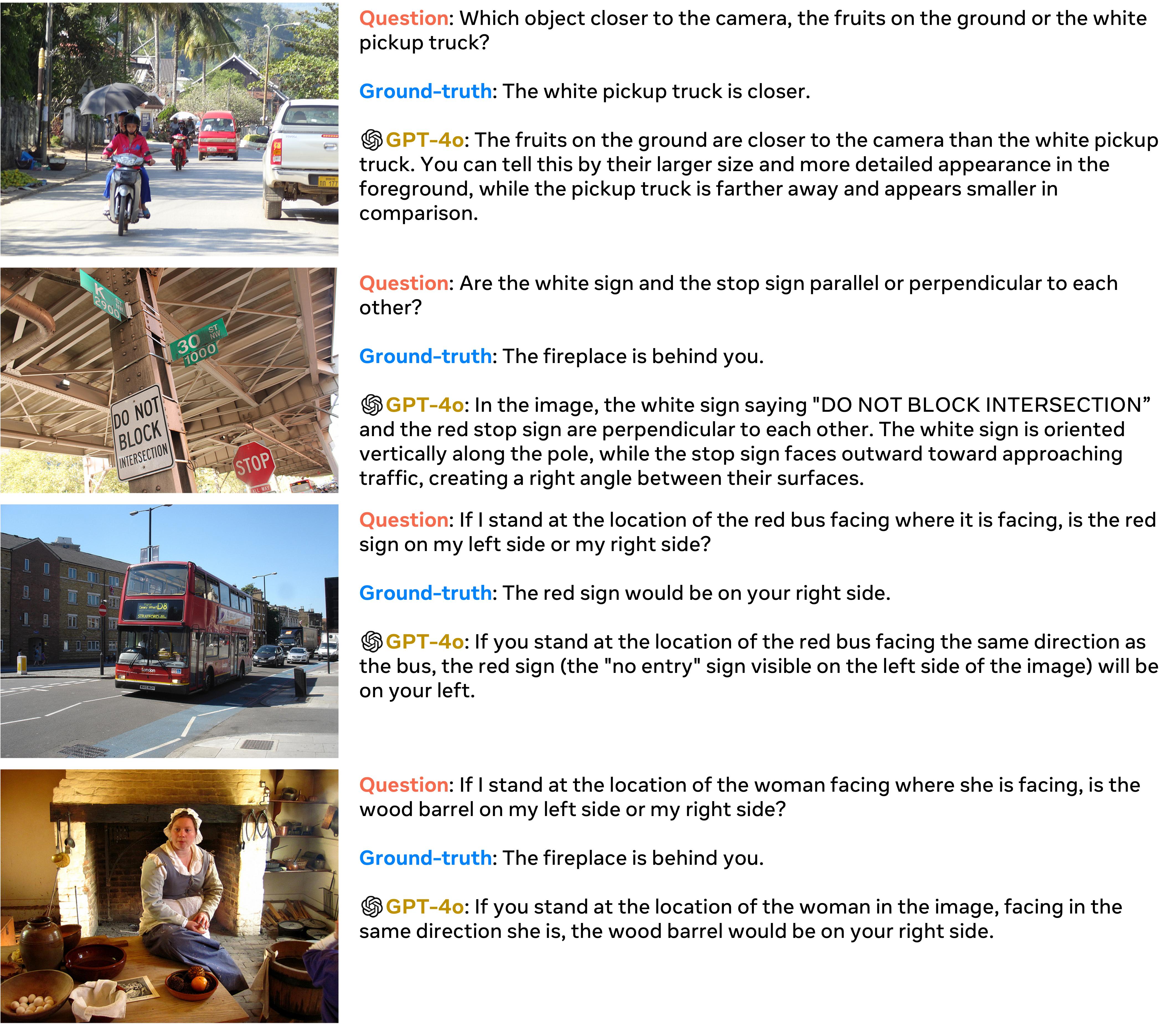}
    \caption{\textbf{Failure cases of GPT-4o~\cite{hurst2024gpt4o} on our \datasetname.} We find that GPT-4o cannot perform rigorous 3D spatial reasoning. In the top figure, GPT-4o resort to visual cues for 3D spatial reasoning. In the bottom two examples, GPT-4o understands the question, attempts to perform 3D spatial reasoning from the scene, and fails to derive the correct answer.}
    \label{fig:supp-failure_cases}
\end{figure*}

\begin{figure*}
    \centering
    \includegraphics[width=\textwidth]{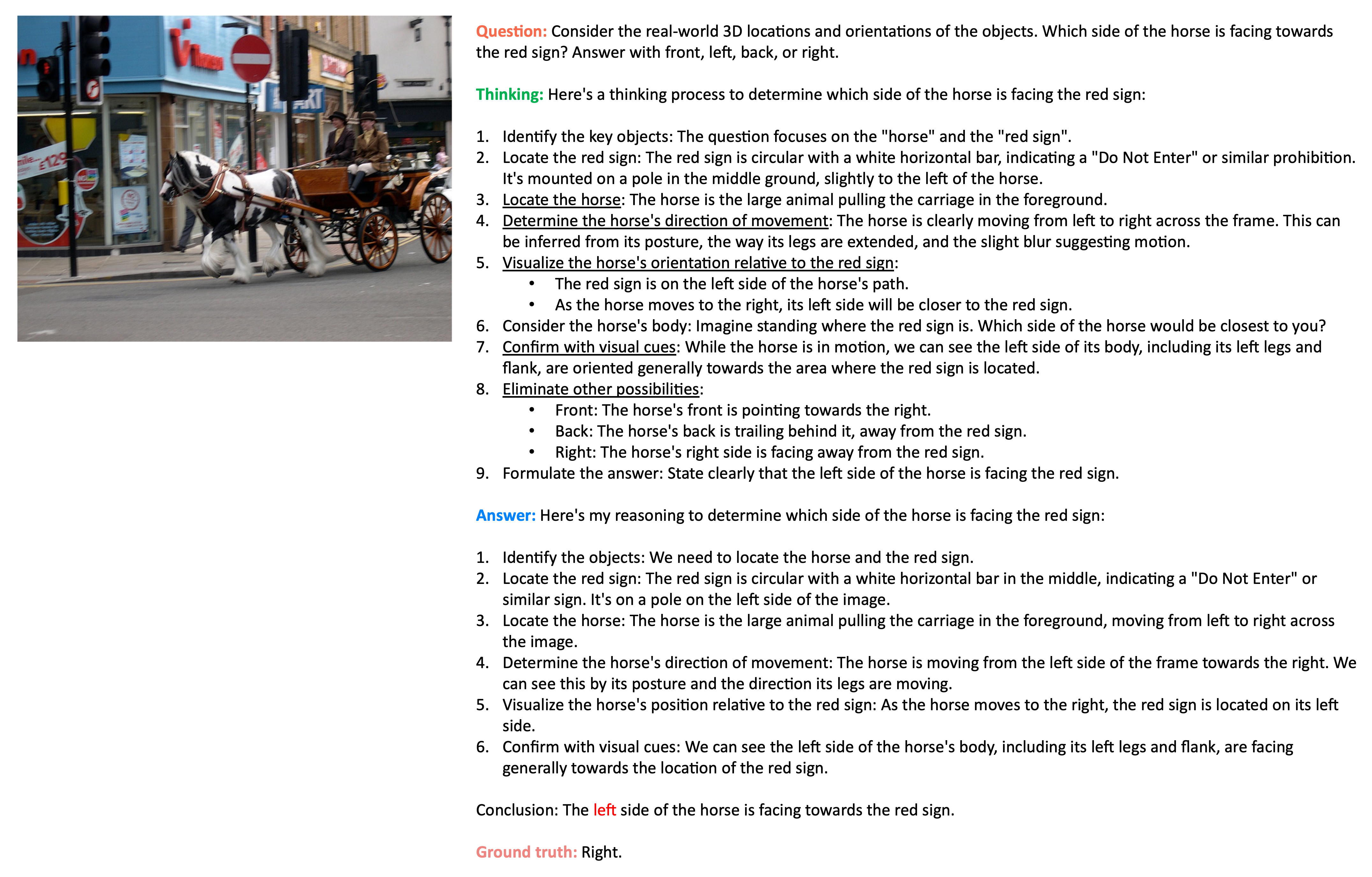}
    \caption{\textbf{Failure cases of Gemini 2.0 Flash thinking~\cite{google2024gemini2} on our \datasetname.} In this example Gemini 2.0 Flash thinking again breaks down the 3D spatial reasoning question into small and tractable steps. It also derives visual cues that can help verify the answer. However, the model still fails to perform 3D spatial reasoning over multiple 3D information, \eg, depth and orientation, and predicts a wrong answer.}
    \label{fig:supp-failure_cases_gemini2}
\end{figure*}

We present some failures cases of GPT-4o~\cite{hurst2024gpt4o} in~\cref{fig:supp-failure_cases} and of Gemini 2.0 Flash thinking~\cite{google2024gemini2} in~\cref{fig:supp-failure_cases_gemini2}.


\end{document}